\documentclass{article}

\usepackage[preprint]{neurips_2026}

\usepackage[utf8]{inputenc} %
\usepackage[T1]{fontenc}    %

\usepackage{adjustbox}
\usepackage{algorithm}
\usepackage{algpseudocode}
\usepackage{amsfonts} %
\usepackage{amsmath}
\usepackage{amsthm}
\usepackage{booktabs}
\usepackage{enumitem}
\usepackage{graphicx}
\usepackage{hyperref} %
\usepackage{microtype} %
\usepackage{nicefrac} %
\usepackage{subcaption}
\usepackage{url} %
\usepackage{xcolor} %

\newtheorem{theorem}{Theorem}
\newtheorem{corollary}[theorem]{Corollary}

\newtheorem{proposition}[theorem]{Proposition}
\theoremstyle{definition}

\newcommand{\Mcal}{\mathcal{M}}
\newcommand{\pev}{p_{\mathrm{ev}}}
\newcommand{\Igz}{I(g;z)}
\newcommand{\Uid}{U_{\mathrm{id}}}
\newcommand{\Gmin}{\Gamma_{\min}}
\newcommand{\Gij}{\Gamma_{ij}}
\newcommand{\Cexec}{C_{\mathrm{exec}}}
\newcommand{\Dub}{\Delta_{\mathrm{ub}}}
\newcommand{\Ureg}{U_{\mathrm{reg}}}
\newcommand{\Dlb}{\Delta_{\mathrm{lb}}}

\title{
Beyond Accuracy: Information-Theoretic Decoder Evaluation Under Latent Intent
}

\author{
Tasha Kim \quad Benjamin Ballyk \quad Anson Lei \quad Ingmar Posner \quad Oiwi Parker Jones \\[0.3em]
University of Oxford \\
\texttt{\{tasha, benjaminb, anson, ingmar, oiwi\}@robots.ox.ac.uk}
}
\begin{document}

\maketitle

\begin{abstract} 
  A central challenge in assistive brain-to-robot control is decoding a user's latent intent from complex and noisy brain signals. Prior approaches evaluate decoders as single-trial classifiers with accuracy-based metrics, but this neglects the inherently sequential nature of assistive control. Since a planner must integrate intermittent, noisy decoder outputs over time, single-step accuracy can only be a proxy when predicting downstream task performance. We formalize brain-to-robot control as a partially observable Markov decision process (POMDP), and prove that there is a sharp phase transition: regret is uniformly bounded across the horizon as long as pairs of goals are all decoder-distinguishable, but grows linearly otherwise. We show that a single channel statistic, the worst-pair Chernoff information $\Gmin$ sets this boundary. Then, we derive a closed-form regret predictor $\Uid(C)$, that can be computed from any confusion matrix in $O(K^2)$ time. Across two domains spanning orthogonal information channels, we show that $\Uid$ predicts cumulative planning regret \emph{beyond what accuracy can explain}. We then highlight a pathological accuracy trap, in which two decoders have identical validation accuracy 35.1\% on real EEG data (BNCI2014001 subject~5) but result in a difference of $9.6\times$ in held-out regret, which $\Uid$ predicts. Same-signal control confirms $\Uid$'s advantage with decoders that span independent feature channels. Ultimately, our work provides a drop-in diagnostic to evaluate and select sensors and decoders in sequential assistive control.
\end{abstract}

\section{Introduction}

Assistive control applications such as brain-to-robot control, gesture-conditioned interaction, often require agents to predict the user's intent based on noisy and ambiguous observations. Recent methodological advances~\citep{defossez2023decoding,jiang2025brainomni} and data curation efforts \citep{mantegna2026libribrain100, schwarz2020analyzing} have made significant progress towards decoders that can accurately infer latent user intent from complex brain and biosignal data from a wide range of sensor modalities. Here, much of the literature formulate intent inference as a supervised classification problem, and evaluate decoder efficacy using prediction accuracy metrics like top-$k$ accuracy and F1 score.
While this paradigm is well-suited to one-shot prediction settings (e.g. brain-to-text prediction \citep{defossez2023decoding, d2025towards}), assistive control has an added layer of complexity as they are inherently \textit{sequential}: decoder outputs are accumulated over time by a downstream planner. As such, standard single-step accuracy evaluations act only as a proxy that may fail to reliably predict downstream task performance. 

We argue that this proxy is fundamentally insufficient.
We model assistive control
as a partially observable Markov decision process (POMDP) \citep{jain2020probabilistic}, in which decoder outputs are treated as observations that inform downstream policies. Through this lens, we directly analyze decoder quality at the task performance level via its induced \textit{cumulative regret}, defined as the gap between the executed policy and an oracle policy with access to the true intent. Using the decoder's confusion matrix, we directly derive information-theoretic bounds on cumulative regret. By formalizing the communication channel between user intent and the policy, we explicitly characterize how decoder quality shapes downstream task performance.

Our central theoretical result gives a sharp \emph{phase transition} on cumulative regret. That is, when all rows of a decoder confusion matrix $C$ are pair-wise distinct,  regret remains uniformly bounded over the horizon $T$. In contrast, regret grows linearly in $T$ whenever any pair of rows collapse, regardless of what the overall accuracy is. This means that overall accuracy does not provide the right measure of performance. Rather, what really matters is the hardest pair of goals to distinguish, which we identify as the minimum Chernoff information, $\Gmin$. We also introduce a computable predictor, $\Uid(C)$ termed the cumulative identification envelope, which can directly estimate regret from the confusion matrix. Together, we show that two decoders with the same accuracy can still incur arbitrarily different regret downstream. Real decoders rarely confuse goals in a symmetric way. For example, in electroencephalogram (EEG) signals, motor imagery (MI) for left and right hands motor imagery are more readily confused than feet and tongue~\citep{pfurtscheller2006mu}.

We validate our theory empirically across multiple datasets and modalities. Through joint regression on the CIFAR-100 dataset that spans 17 decoders and 7 feature families, we show that $log(\Uid)$ explains around $89\%$ of the regret variance. Similarly on the NinaPro multimodal dataset, $log(\Uid)$ carried the signal, while adding accuracy did not make a significant difference. When the accuracy of two decoders are closely matched on the aforementioned datasets, $\Uid$ correctly predicts which decoder will incur the higher regret in ${\sim}77\%$ instances. On a real EEG decoder pair (BNCI2014001~\citep{tangermann2012review} subject 5) that has identical validation accuracy (35.1\%), held-out planning regret gave a $9.6\times$ difference, and $\Uid$ spots the right one. Finally, three same-signal negative control experiments with decoders using single-feature representation tests the theory's predicted boundary. From this, we identify that accuracy captures most of the variation that is relevant to downstream performance when channels are not orthogonal. Despite this, $\Uid$ outperforms precisely when channels are genuinely orthogonal, matching our theoretical results.

\section{Notations and background}
\label{sec:notations}

\textbf{Environment and costs.} Consider a finite-horizon shared-autonomy problem in which, at each discrete time step
$t \in \{1,\ldots,T\}$, a robot occupies a state $x_t \in \mathcal{X} \subseteq \mathbb{R}^d$
and executes an action $a_t \in \mathcal{A}$. Here, $\mathcal{X}$ denotes the robot state
space, $\mathcal{A}$ the action space, and $d$ the dimension of the state. $T$ is the total number of control steps in the horizon. At the beginning of each episode,
a true goal $g^* \in [K] := \{1,\ldots,K\}$ is sampled uniformly,
$g^* \sim \mathrm{Unif}([K])$, where $K$ is the number of possible reachable goals. Each goal
$g \in [K]$ is associated with a target state $x^*(g) \in \mathcal{X}$. The per-step cost
incurred at time $t$ under $g^*$ is
$$
c_t = c(x_t,a_t;g^*) =
\|x_t-x^*(g^*)\|_2 + \lambda \|a_t\|_2^2,
$$
where the first term penalizes distance to the target state for the true goal, the second
term penalizes control effort, and $\lambda \geq 0$ is a weighting parameter. All experiments set $\lambda=0$, so the implemented cost is distance to target only. We compare a planner policy $\pi$ against an oracle that observes $g^*$ and therefore follows the
optimal goal-conditioned policy $\pi^*_{g^*}$. Regret measures how worse a planner is compared to when the robot knows its goal. We define the cumulative regret of $\pi$
over horizon $T$ as
$$
R_\pi(T)
:=
\sum_{t=1}^{T}
\mathbb{E}\!\left[
c_t^{\mathrm{plan}} - c_t^{\mathrm{oracle}}
\right],
$$
where $c_t^{\mathrm{plan}}$ is the cost incurred by the planner at time $t$,
$c_t^{\mathrm{oracle}}$ is the cost incurred by the oracle at time $t$, and the expectation
is taken over the random goal, any stochasticity in the planner, and any induced trajectory
randomness.

\textbf{Decoder channel.} At each time $t$, the planner receives a noisy decoder output $z_t \in [K]$ with probability $p_{ev}$, via the confusion matrix $P(z_t = j \mid g^* = i) = C_{ij}$. The sum of each row in the confusion matrix equals $1$, hence $\sum_{j}C_{ij}=1$. It receives $\bot$ when no observation arrives. Hence, each goal $i$ induces an erasure-augmented row distribution $C_{ij}$ on $\{\bot, 1, \cdots, K\} $ with $P(y_t = \bot \mid g^* = i) = 1 - \pev$ and $P(y_t = j \mid g^* = i) = \pev C_{ij}$.

\textbf{Information quantities.} We use information-theoretic quantities to characterize how well an intermittent decoder infers a user's latent intent. The one-step mutual information of the intermittent decoder is $I(g^*; y_t) = \pev \cdot \Igz$. For $i \neq j$, the pairwise Chernoff information of the erasure-augmented rows is
\begin{equation}
\Gij := -\log \min_{s \in [0,1]} \!\left[ (1 - \pev) + \pev \sum_{m=1}^{K} C_{im}^{1-s} C_{jm}^{s} \right], \qquad \Gmin := \min_{i \neq j} \Gij.
\label{eq:gamma_def}
\end{equation}

The identification envelope is
\begin{equation}
\Uid(C) := \frac{1}{K} \sum_{i=1}^{K} \sum_{j \neq i} \frac{1}{e^{\Gij} - 1},
\label{eq:Uid_def}
\end{equation}

which is finite whenever all rows of $C$ are distinct. The Fano converse below uses global information $\pev \Igz$, while the achievability bound is governed by $\Gmin$. For symmetric channels these are monotone in the same scalar accuracy parameter, but for asymmetric channels they can disagree sharply. This asymmetry drives our central prediction that \emph{accuracy is not the right planning metric}.

\textbf{Planner.} Our identification bounds are agnostic to policy. Hence, we need to link misidentification and excess cost to translate them to regret. We focus on \emph{goal-committing} planners: at each time $t$, the planner selects a label $\tilde{g}_t \in [K]$ measurable with respect to its information and executes the oracle policy $\pi^*_{\tilde{g}_t}$. This class of planners admits an exact bounded-versus-linear regret phase transition without auxiliary assumptions, captures the standard maximum a posteriori (MAP) pipeline used in brain-computer interfaces (BCIs) and shared-autonomy settings, and matches the planner used in all our experiments. The more general case of hedge-capable planners is discussed in Appendix~\ref{app:hedge}.

\section{Theoretical results}
In this section, we present the main theoretical results. We derive bounds on cumulative regret that links single-step decoder characteristics to controller performance in two steps: first, we bound the cumulative identification error, $\Mcal(T) := \sum_{t=1}^{T} P(\hat g_t\neq g^*)$, using only the decoder's channel statistics (Theorems~\ref{thm:lower}-~\ref{thm:upper}). Here, $\hat g_t$ denotes the planner's inferred goal after receiving an observation at time $t$. Second, we translate this identification error bound to an exact two-sided bound on regret for goal-committing planners (Proposition~\ref{prop:commit}). This allows us to link the statistical properties of the decoder output to task performance, and a sharp phase transition emerges as a corollary (Corollary 4).

Our results rely on the following assumptions:
\begin{itemize}[label={}]
    \item \textbf{A1.} Decoder observations are \emph{action-independent}: $P(y_t \mid g^*, x_t, a_t) = P(y_t \mid g^*)$.
    \item \textbf{A2.} Evidence is independent, so $y_t | g^*$ are \emph{independent and identically distributed} (i.i.d).
    \item \textbf{A3.} Costs are \emph{bounded}: $0 \leq c_t \leq c_{max}$.
    \item \textbf{A4.}  Channels $C$ are known (theorem statements) or estimated  on a calibration split that is disjoint from the evaluation (experiments).
\end{itemize}
These assumptions reflect a common assistive control scenario where the user's goal is fixed across each episode, and a single decoder with stable channel characteristics is repeatedly applied to infer it. With this abstraction, we now show how identification error can be bounded using decoder statistics.

\subsection{Bounding Identification Error}

The following theorems provide a lower and upper bounds to the cumulative identification error.

\begin{theorem}[Lower regret bound]
\label{thm:lower}
Under A1-A4, for any estimator $\hat{g}_t$ that can be measured with respect to the planner's information at time $t\in [T]$, 
$$P(\hat g_t\neq g^*) \ge \left[1-\frac{t\,\pev\,\Igz + \log 2}{\log K}\right]_+.$$

Subsequently, \begin{equation}
  \Mcal(T)
  \ge
  \sum_{t=1}^{T}
  \left[
    1-\frac{t\,\pev\,\Igz + \log 2}{\log K}
  \right]_+.
  \label{eq:M_lower}
\end{equation}
\end{theorem}

\emph{Proof sketch.} \textbf{A1} gives $g^*\rightarrow y_{1:t}\rightarrow\mathcal{D}_t$. $I(g^*; y_{1:t})$ is less than or equal to $t\cdot \pev I(g;z)$ from \textbf{A2} and erasure. Then Fano on $\mathcal{D}_t$ gives per-step bound. Full proof is included in the Appendix~\ref{app:lower}.

Intuitively, this means that information about $g^*$ accumulates linearly at most. Fano's inequality then converts this information bound into a lower bound on the probability of the error of any estimator.

Next, we use a Chernoff argument to establish an upper bound on the MAP rule, which is a natural choice of discrete classifier that is used for assistive control.

\begin{theorem}[Upper regret bound]
\label{thm:upper} 
Under A1-A4 with a uniform prior, the MAP estimator satisfies 
\begin{equation}
\begin{split}
  P(\hat g_t^{\mathrm{MAP}}\neq g^*) 
    &\le \frac{1}{K}\sum_{i=1}^{K}\sum_{j\neq i} e^{-t\Gij} \\
    &\le (K-1)\,e^{-t\Gmin}.
\end{split}
\end{equation}
And consequently, 
\begin{equation}
\begin{split}
\Mcal_{\mathrm{MAP}}(T) 
&\leq \Uid(C) \\
&\leq \frac{K-1}{e^{\Gmin}-1}.
\end{split}
\end{equation}

Whenever all rows of the channel $C$ are distinct, $\Mcal_{\mathrm{MAP}}(T)$ is bounded uniformly in $T$.
\end{theorem}

\emph{Proof sketch.} Chernoff bounds gives $e^{-t\Gij}$  per pair. Union bound and geometric summation gives $\Uid(C)$. Full proof is included in the Appendix~\ref{app:upper}.

Our result holds for any MAP estimators i.e., decoders that output a single class label per trial. This matches the standard assistive control setting in which intent decoders are classifiers that produce a single discrete prediction. 

Theorem~\ref{thm:upper} establishes the theoretical underpinning of our analysis. Firstly, the theorem gives a concrete, computable upper-bound to the identification error, $\Uid$. This serves as a predictor for downstream task performance. Additionally, the theorem illustrates that cumulative identification error is governed not only by aggregating classification accuracy, but also by the worst-case distinguishability between goals, captured by $\Gmin$. Intuitively, this means that the rate at which errors are accumulated is controlled by how \textit{confusable} goals are with respect to the decoder.

\emph{Why converse and achievability use different quantities.} $I(g;z)$ controls aggregate uncertainty reduction. The MAP achievability is controlled by the \emph{worst confusable pair} ($\Gmin$), because the overall error event is a union over pairwise confusions. For symmetric channels these reduce to monotone functions of accuracy and agree, but for real asymmetric channels they can disagree sharply. This is a structural gap that enables our central prediction that accuracy is not the right planning metric, which we test empirically in Section~\ref{sec:experiments}.

\subsection{From identification to regret}

Identification error translates exactly to regret in goal-committing planners. We define the per-step wrong-goal gaps $\delta_{ij} := \mathbb{E}[c_t^{\pi^*_j} - c_t^{\pi^*_i} \mid g^* = i]$, $i \neq j$, and assume:

\begin{itemize}[label={}]
    \item \textbf{A5-C.} There exist some $\Dlb$ and $\Dub$ such that $0 < \Dlb \leq \Dub < \infty$ where $\Dlb \leq \delta_{ij} \leq \Dub$ for all $i \neq j$.
\end{itemize}

This means that there is a per-step cost penalty to committing to the wrong goal, and this cost is bounded by constants above and below. This assumption holds for any environment that has bounded and well-separated goal-conditioned policies.

\begin{proposition}[]
\label{prop:commit}
Under A1-A4 and A5-C, if a planner commits to $\tilde{g}_t = \hat{g}_t^{\mathrm{MAP}}$ at each step, then
\begin{equation}
\begin{aligned}
\Dlb \cdot \Mcal_{\mathrm{MAP}}(T)
&\;\leq\; R(T) \\
&\;\leq\; \Dub \cdot \Mcal_{\mathrm{MAP}}(T) \\
&\;\leq\; \Dub \cdot \Uid(C) \\
&\;\leq\; \Dub \cdot \frac{K-1}{e^{\Gmin}-1}.
\end{aligned}
\label{eq:regret-sandwich}
\end{equation}
\end{proposition}

The full proof is in Appendix~\ref{app:commit}. 
In our experiments, we use $\Ureg(\hat{C}) := \Dub \cdot \Uid(\hat{C})$ as the predictor for regret bound.  $\Dub$ here is an environment-level constant (see Section~\ref{sec:4-4}). We use a standard reaching setup described in Section~\ref{sec:4-1} throughout, with $\Dub$ value set to $2.0$. Analogous results for hedge-capable planners are provided in Appendix~\ref{app:hedge}.

\subsection{The phase transition}
\begin{corollary}[Exact phase transition on regret]
\label{cor:phase-transition} Under A1-A4, A5-C and MAP-commit planner, the following are true:
\begin{enumerate}
    \item If rows of confusion matrix $C$ are all distinct, then $\Gmin > 0$, and the cumulative MAP misidentification is bounded (by Theorem 2). 
    \item If, for some $i \neq j$, $C_i = C_j$, then for every estimator $\hat{g_t}$
    , $P(\hat g_t\neq g^*) \geq \frac{1}{K}$ and $R(T)$ is greater than or equal to $\Delta_{lb} \cdot \frac{T}{K}$. 
\end{enumerate}

\emph{Proof.} Combining Proposition~\ref{prop:commit} and Theorem~\ref{thm:upper} gives 1. For 2, $C_i = C_j\Rightarrow\widetilde{C}_i = \widetilde{C}_j$, hence no estimator can distinguish $i$ from $j$ with average error $<0.5$ on the pair. Combining with the uniform prior, we get $P(\hat{g}_t \neq g^*) \geq 1/K$. Proposition~\ref{prop:commit} turns this into linear-in-$T$ regret lower bound. \qed

\end{corollary}

Notably, we see that regret transitions \emph{exactly} from $O(1)$ to $\Omega(T)$ at the boundary $\Gmin = 0$, regardless of the overall accuracy. Our corollary exposes a pathological oversight: a decoder achieving 90\% accuracy but $\Gmin = 0$ is fundamentally unsuitable for sequential planning, as its cumulative identification error is unbounded over time. In contrast, a decoder with a significantly lower accuracy of $50\%$ but with $\Gmin > 0$ may saturate to bounded regret over time.

Consequently, this corollary highlights a failure mode that remains invisible to aggregate accuracy metrics. 
In Section~\ref{sec:experiments}, we show near-degenerate cases of $\Gmin \approx 0$ found in real-world datasets, demonstrating its empirical relevance, rather than being a mere theoretical artifact.

\begin{figure}[t]
  \centering
  \includegraphics[width=\columnwidth]{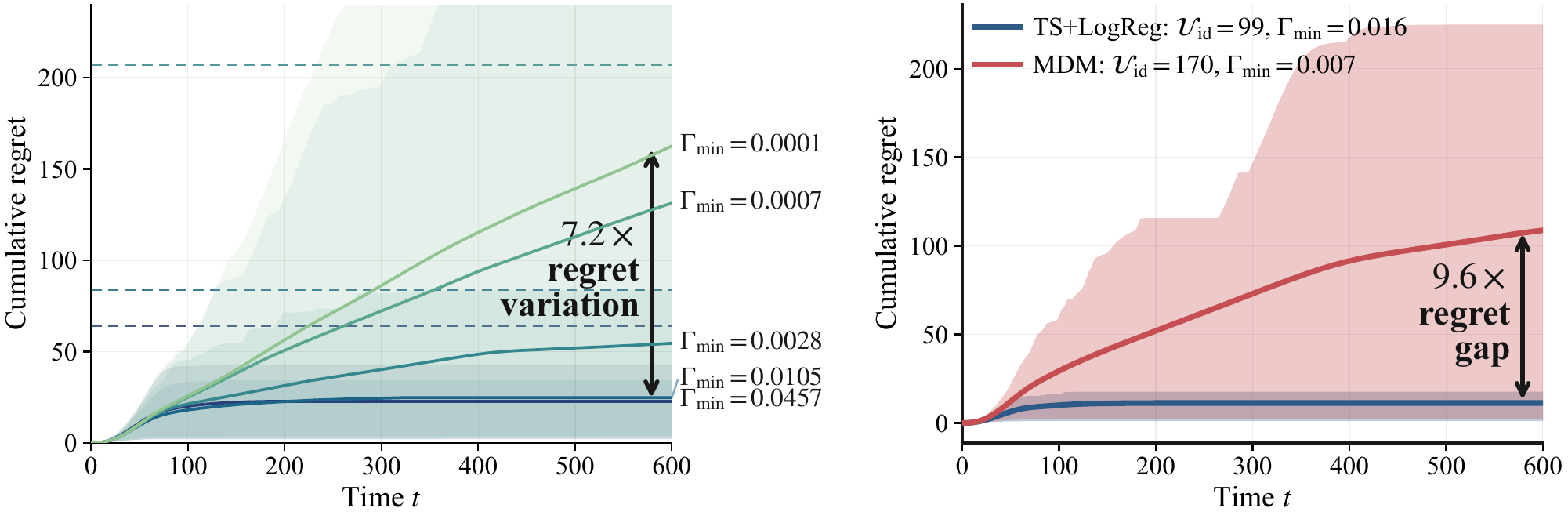}
  \caption{ \textbf{Decoder distinguishability governs planning regret}. (\emph{Left}): 
  Five confusion-matrix channels with \emph{identical} decoder accuracy ($\alpha{=}0.50$, $K{=}4$). Varying 
  $\Gmin$ produces $7.2\!\times$ empirical cumulative regret ($22.7$$\rightarrow$$162.6$) at $T{=}600$. 
  \emph{Dashed lines:} theoretical envelope $\Dub\Uid(C)$.
  (\emph{Right}):  Real EEG, illustrative (BNCI2014001 Subject~5): two decoders with \emph{identical} validation accuracy ($20/57$ trials), but $\Uid{=}99.4$ vs. $170.5$. Held-out replay (accuracy $48.3\%$ vs.\ $34.7\%$) gives $11.3$ vs.\ $108.8$ regret at $T{=}600$, as $\Uid$ predicts.
}
\label{fig:headline}
\vspace{-1em}
\end{figure}

\vspace{-0.2em}
\section{Experiments}
\label{sec:experiments}

We test the following claims: (C1) $\Uid$ predicts the cumulative planning regret $beyond$ what accuracy alone explains, when decoders span multiple orthogonal information channels (Section~\ref{sec:4-2}). (C2) The predicted boundaries hold, and the advantage of $\Uid$ disappears on same-signal decoder families (Section~\ref{sec:4-3}). (C3) The theoretical regret envelope $\Delta_{ub}\cdot \Uid$ holds on held-out test data (Section~\ref{sec:4-4}). We prevent post-hoc fitting by estimating $\Gij, \Gmin, \Uid$ on validation splits that are disjoint from regret evaluation. Additionally, $\Delta_{ub}=2.0$ is the per-step cost-gap determined \emph{beforehand}, only from the environment geometry (see 
Algorithm~\ref{alg:protocol}
of Appendix~\ref{app:protocol}). Importantly, we do not tune, filter, nor calibrate the cohort for $\Uid$ at evaluation-time.

\subsection{Experimental setup}
\label{sec:4-1}
\begin{figure}
  \centering
  \includegraphics[width=\columnwidth]{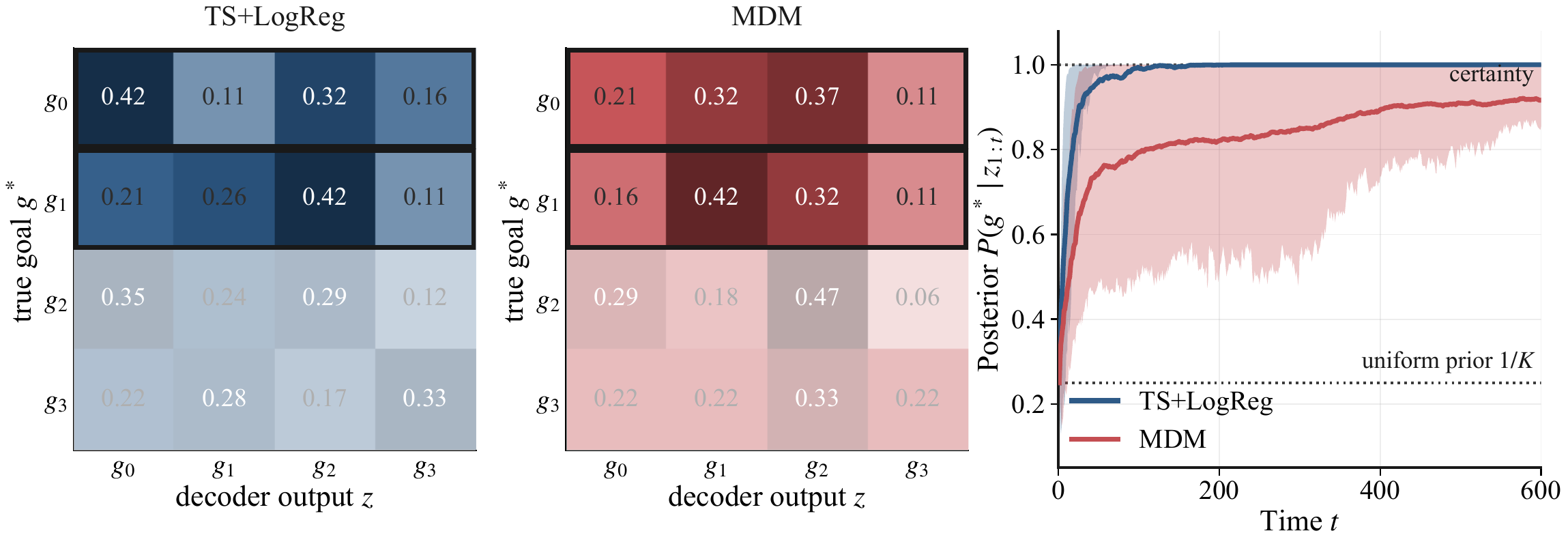}
  \caption{\textbf{Accuracy is the wrong objective when a planner must identify the goal.}  (\emph{Left}): TS+LogReg: two row distributions stay well-separated, $\Gamma_{\min}=0.016$, $\mathcal{U}_{\mathrm{id}}=99$. (\emph{Middle}): MDM: two rows markedly \emph{closer}, so each observation only weakly shifts posterior, %
  and the planner needs many more samples to disambiguate,  $\Gamma_{\min}=0.007$, $\mathcal{U}_{\mathrm{id}}=170$.
  (\emph{Right}): Posterior on the true goal over $n=120$ episodes (shaded: $16$--$84$th percentile). Two decoders achieve identical validation accuracy ($35.1\%$), but held-out planning regret diverges by $\mathbf{9.6\times}$ on BNCI2014001 subject 5.
  }
\vspace{-0.8em}
\label{fig:acc-trap}
\end{figure}

\textbf{Datasets.}
\begin{enumerate}[itemsep=0pt, topsep=2pt]
\item For bounds checking, we use two BCI datasets: (i) Stieger2021~\cite{stieger2021benefits} contains $4$-class human electroencephalogram (EEG) motor imagery data recorded from $62$ subjects ($61$ with usable sessions, $15$ calibration and $46$ held-out), with $1{,}600$ held-out conditions ($1{,}594$ with finite $\Uid$). (ii) BNCI2014001~\cite{tangermann2012review} is a $4$-class EEG motor imagery dataset collected from $9$ subjects. We use the latter for external replication, with no recalibration efforts.  
\item For feature family and cross-modality decorrelation, we use (iii) CIFAR-100~\cite{krizhevsky2009learning} as a benchmark that spans 7 feature families across 60K images grouped into 100 classes and 20 superclasses, capturing genuinely independent image properties like color, spatial layout, texture, and shape. (iv) NinaPro DB2~\cite{atzori2014electromyography} is a multimodal dataset including 4 physical modalities. 
\item For same-signal controls, we use covariance-feature decoders (CSP+LDA and three Riemannian) from Stieger2021, MFCC-only from Google Speech Commands~\cite{speechcommands2018}, and EMG-only from NinaPro.
\end{enumerate}

\vspace{0.2em}
\textbf{Planners.} We implement a canonical, 2D double-integrator reaching environment. There are K goals on a unit circle that are equally spaced. We use parameters $\text{damping}=0.1$, $\text{gain}=2.0$, and dt$=0.05$, with a goal-committing (MAP) planner, and observe \emph{how regret evolves across the horizon}.  We vary the decoder channel $C$ across tested domains, where the shared environment is able to isolate the exact effects of the confusion-matrix channel. Note $T=600, 300, 300$, and $p_{ev} = 1.0, 0.2, 0.5$ were used for (i)+(ii), (iii), and (iv) datasets, respectively. 

\subsection{(C1) $\Uid$ predicts regret beyond accuracy}
\label{sec:4-2}

In this section, we test whether $\Uid$ better correlates to actual cumulative regret compared to accuracy. First, we regress cumulative regret on $\alpha$ alone, second $\log(\Uid)$\footnote{We use $\log(\Uid)$ here instead of $U_{reg} = \Delta_{ub}\cdot \Uid$, since $\Delta_{ub}$ scales but does not change rank order. Additionally, log-linear regression gives the appropriate model for rate-of-decay quantity that $\Uid$ encodes.} alone, and finally jointly. Table~\ref{tab:regression} summarizes our results. We report $R^2$ for each single-predictor regression, improvement in the joint-model $\Delta R^2$, and partial $F$-tests if each predictor contributes unique variance given the other.

Our results reveal a pattern precisely aligning with what our theory predicts. In (I) CIFAR-100, $\Uid$ explains $89\%$ of regret variance. In fact, accuracy seems to add nothing beyond what $\Uid$ provides in the joint model. The same thing holds for (II) NinaPro multimodal, where accuracy adds no significant unique variance beyond $\Uid$.  Finally, NinaPro EMG-only highlights a negative result in decoders that share a single signal channel with a caveat. The confusion matrices are built from 35 test trials spread over 23 classes, and smoothing leaves many rows equal, resulting in $338/440$ conditions having $\Gmin{=}0$ and $\Uid{=}\infty$. Even so, ranking all 440 conditions shows accuracy is a better predictor ($\rho(\alpha,R)=-0.748$ vs.\ $\rho(\Uid,R)=+0.305$), which corroborates what our theory states about what should happen when every decoder sees the same signal. On the 102 resolvable conditions alone, the result reverses ($R^2_\alpha=0.356$ vs.\ $R^2_{\log\Uid}=0.450$, a per-subject direction accuracy of $68.3\%$ over the 89 matched pairs), but this subset does not include most weak decoders so we report this without relying on it. The same-signal control with real power is shown in Stieger2021 (Section~\ref{sec:4-3}).

\textbf{Matched-accuracy direction tests.} The regression results establish that $\Uid$ adds variance over accuracy in expectation. We also test a sharper version of the claim: at equivalent accuracy, can $\Uid$ identify which decoder will incur higher regret? 
$\Uid$ correctly predicted the regret direction in $76.9\%$ of 39 pairs with matched accuracy on CIFAR-100. 70 within-subject cross-modality pairs on NinaPro multimodal also gave a per-subject mean of $77.0\%$ ($\Uid=\infty$ excluded conditions).
See Appendix~\ref{app:ninapro-persubject} for per-subject details.

\textbf{The accuracy trap.} Figure~\ref{fig:headline} \emph{(Right)} and Figure~\ref{fig:acc-trap} demonstrate this failure mode in real EEG data as a concrete example on BNCI2014001. For subject 5, tangent-space logistic regression (TS+LogReg) and minimum distance to mean (MDM) decoders both yield $35.1\%$ validation accuracy. In contrast, a replay of their held-out channels shows the commit planner incurring $9.6\times$ higher regret with MDM than with TS+LogReg. $\Uid$ saw this coming, however, giving $170.5$ for MDM against $99.4$ for TS+LogReg. Taking a closer look, we see the MDM's confusion matrix has a class pair that is difficult to tell apart, so each observation hardly moves the posterior between the two goals. Compared to this, pairs in TS+LogReg stay cleanly separated. The theory successfully identifies the gap where accuracy as a metric is blind. In particular, two decoders may score similar accuracy, but which pair(s) of goals a decoder confuses decide the quality of a planner over time. We let this pair serve as an illustration, given only around $14$ validation trials per class. Note Figure~\ref{fig:acc-trap} shows Laplace-smoothed matrices.
The existence claim is shown as a controlled experiment in Figure~\ref{fig:headline} \emph{(Left)}.

\begin{table}[t]
\centering
\caption{
\textbf{$\mathbf{\Uid}$ vs. accuracy as predictors of cumulative regret for decoders spanning orthogonal information channels.} $R^2$ from linear regressions of cumulative regret on accuracy $\alpha$, $\log \Uid$, and their joint model. $n$ counts conditions with finite $\Uid$.
$\Delta R^2_{\mathrm{joint}}$ is the improvement of the joint model relative to accuracy alone. Confidence intervals are bootstrap $95\%$ CIs on $R^2_{\log (\Uid)} - R^2_{\alpha}$ over conditions. Partial-$F$ tests evaluate whether each predictor explains unique variance given the other.
$\Uid$ predicts regret beyond what accuracy explains.
}
\label{tab:regression}
\vspace{1.2em}
\setlength{\tabcolsep}{5pt}
\begin{adjustbox}{width=\columnwidth}
\begin{tabular}{lllllcl}
\toprule
Domain 
& $n$ 
& $R^2_{\alpha}$ 
& $R^2_{\log \Uid}$ 
& $\Delta R^2_{\mathrm{joint}}$ 
& 95\% CI 
& Partial-$F$ result \\
\midrule
(I) CIFAR-100 (7 families)
& 17
& 0.702
& \textbf{0.894}
& $+0.192$
& $[+0.05,\,+0.34]$
& $\Uid$ subsumes $\alpha$ \\
(II) NinaPro multimodal (4 modalities)
& 130
& 0.659
& \textbf{0.755}
& $+0.097$
& $[+0.04,\,+0.16]$
& $\Uid$ subsumes $\alpha$ \\
\bottomrule
\end{tabular}
\end{adjustbox}
\end{table}

\begin{table}[t]
\centering
\caption{
\textbf{Same-signal negative controls.} When all decoders operate on a single feature channel, accuracy $\alpha$ and $\Gmin$ become tightly coupled ($r \geq 0.80$ as seen across three independent domains), and $\Uid$ provides little or no additive predictive signal. This is consistent with the theoretical prediction, and confirms $\Uid$'s utility in multi-channel settings: an advantage over accuracy that is tied specifically to decoder families that span orthogonal information channels.
}
\label{tab:negative-controls}
\vspace{1.2em}
\small
\setlength{\tabcolsep}{4pt}
\renewcommand{\arraystretch}{0.95}
\begin{tabular}{l l l c}
\toprule
Domain 
& Feature channel 
& $n$ 
& $r(\alpha,\Gmin)$ \\
\midrule
Motor imagery (Stieger2021)
& Covariance features
& 1{,}600
& 0.80 \\
Speech Commands
& MFCC features
& 11 decoders
& 0.96 \\
NinaPro EMG-only
& Time-domain EMG
& 11 decoders
& 0.82 \\
\bottomrule
\end{tabular}
\vspace{-0.8em}
\end{table}

\subsection{(C2) The boundary: orthogonal channels are required}
\label{sec:4-3}
The regression and direction test results reveal a unique condition for which our theory empirically provides value. Decoders that share single feature representations have confusion matrices that are structurally very similar. However, $\Uid$ proves real advantage in  domains where decoders span independent feature channels  such as image color, texture, gradients in CIFAR-100, and in NinaPro multimodal. In these cases, $\Uid$ captures this pair-specific structure. In real-world human-robot interaction and shared autonomy contexts that often combine a diverse range of sensors such as a robot stack that utilizes eye-gaze, EMG, IMU and speech for communication and control, $\Uid$ can give a distinct and pragmatic advantage. See Table~\ref{tab:negative-controls} for three same-signal control domains. 
Stieger2021 shows the strongest same-signal control, as $\Uid$ spots the decoder with higher regret in a per-subject mean of only $56.2\%$, over the 417 pairs with matched accuracy. Only $24/46$ subjects are above chance, showing that $\Uid$ is near chance when decoders share a feature channel.

\subsection{(C3) Theoretical envelope holds on real data}
\label{sec:4-4}

Beyond predicting and ranking regret, $\Uid$ also provides a quantitative upper bound on regret (Proposition~\ref{prop:commit}), $R(T) \leq \Dub \cdot \Uid(C)$. Note in our canonical reaching task environment, the value $\Delta_{ub}$ is determined only through the goal geometry without needing per-subject calibration. That is, $K=4$ goals are on the unit circle, and the opposite goal's asymptotic excess cost is $2$. We verify this envelope on both of the BCI datasets.

On BNCI2014001, the bound holds on $100\%$ of held-out conditions ($36/36$) with a median bound-tightness value of $\hat R(T)/\Ureg = 0.13$. On the Stieger2021 dataset, bound holds on $85.8\%$ ($1{,}367/1{,}594$) of the held-out conditions. Note that Proposition~\ref{prop:commit} bounds expected regret, while we measure empirical $\hat R(T)$  from $120$ episodes for each condition. 

We observe that violations cluster at conditions where $\Gmin$ is large, but $\Uid$ is sufficiently small as to allow a mis-commit on a single episode push $\hat R(T)$ above the small absolute bound while $\mathbb{E}\,[\,R(T)]$ stays below it. The bound remains informative as a deployment-readiness check. The predictive case of $\Uid$ as a decoder-evaluation metric is supported by the regression evidence in Section~\ref{sec:4-2}. 

\begin{figure}
  \centering
\includegraphics[width=\columnwidth]{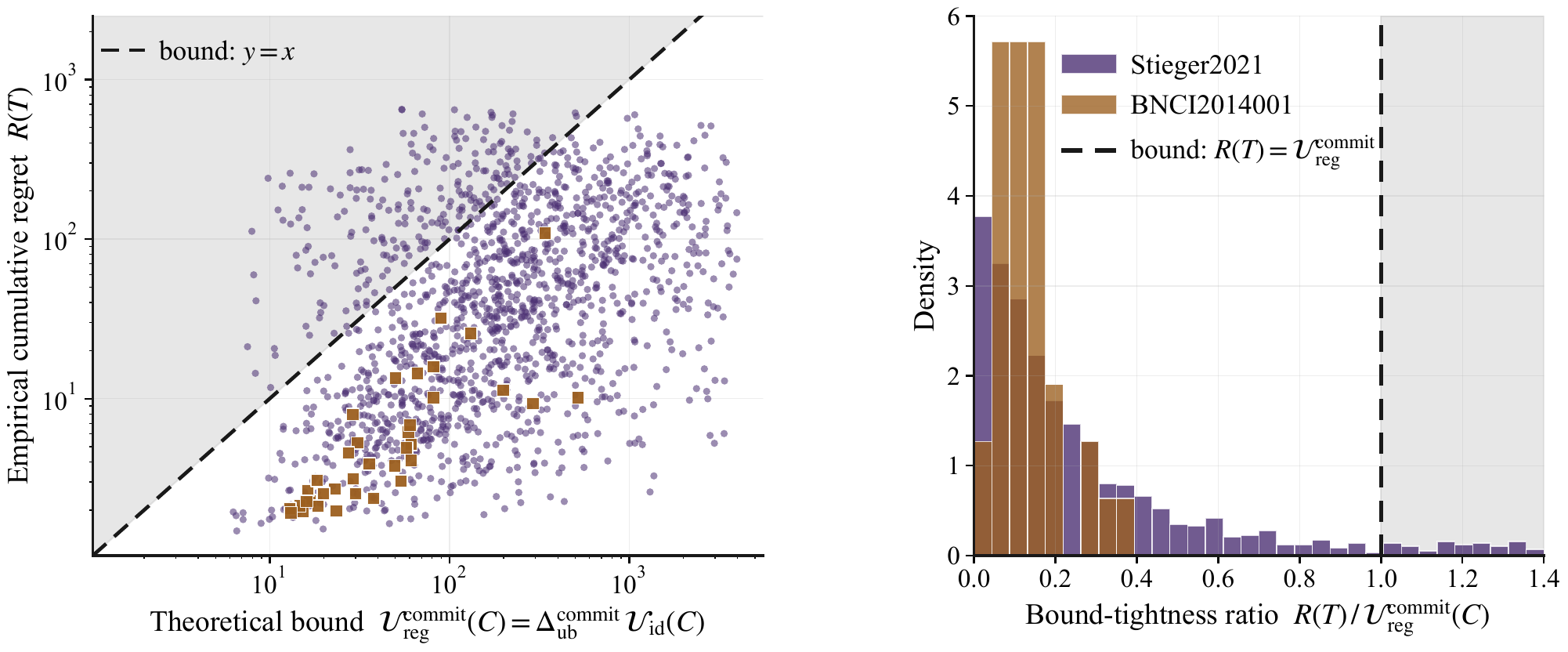}
\vspace{-0.4em}
\caption{\textbf{Empirical verification of the regret envelope.} Each point is a held-out evaluation condition, and shaded grey region marks bound violations. (\emph{Left}): Empirical cumulative regret $R(T)$ vs. the commit-bound prediction $\mathcal{U}_{\mathrm{reg}}^{\mathrm{commit}}(C) = 2\,\mathcal{U}_{\mathrm{id}}(C)$ for both datasets (log-log). (\emph{Right}): Distribution of bound-tightness ratio $R(T) / \mathcal{U}_{\mathrm{reg}}^{\mathrm{commit}}(C)$. External replication on BNCI2014001 (36 conditions across 9 subjects) shows 36/36 satisfy the bound, median tightness $0.13$. On Stieger2021 ($1{,}594$ conditions, 46 held-out subjects), $85.8\%$ satisfy the bound. }
\vspace{-1em}
\end{figure}

\vspace{-0.2em}
\section{Related work} 
We review prior work relevant to our setting. Our perspective across these fields is planner-centric: we ask which statistical property of the decoder's channel governs regret over a long horizon.

\paragraph{Assistive brain-robot interfaces and evidence accumulation.}  In assistive robotics, neural-signal operated systems such as NOIR~\citep{zhang2023noir,kim2025noir2, zhang2025eegbci} have helped to augment or restore users'  daily communication and control capabilities \citep{millan2004noninvasive}.
In BCI and neural decoding applications, both classical and deep learning-based decoders are still largely evaluated through single-trial accuracy or rate of information transfer \citep{ramoser2000optimal}. 
Sequential pipelines such as these, however, must rely on the accumulation of evidence over time or on dynamic stopping, due to the inherent noisy nature of neural signal measurements \citep{mainsah2015increasing}. Hence, calibrating and evaluating decoders remain a crucial step before deployment, in order to ensure safety and downstream task performance \citep{lotte2018review}. Our work explores the case where aggregate accuracy of a decoder that feeds a downstream planner cannot act as a sufficiently reliable proxy in performance over time.

\paragraph{Noisy decoders and latent-goal inference.} Assistive or shared autonomy systems have formulated this problem as planning under latent goal uncertainty. In this setup, a robot or planner must actively act or disambiguate noisy observations~\cite{sadigh2016information}, while simultaneously inferring the user's intent. Previous works have formalized this shared control problem through policy blending \citep{dragan2013policy}, or planning in the belief space, such as a POMDP-style formulation over a hidden goal, or Bayesian intent inference from observations that make the latent-goal structure more explicit \citep{javdani2018shared,jain2020probabilistic}. These works highlight that  \emph{informativeness} of human signals matter as much as raw accuracy in human-robot interactive systems. Biological signals could be present in the form of eye-gaze, gestures, language, electromyography (EMG), or brain signals, all of which give useful and distinct information about a user's goal or intent \citep{zhang2019eeg}. We complement this literature by characterizing the channel statistics that matter for evaluating how much performance a robot or planner loses, as it makes decisions over a long horizon.

\paragraph{Information-theoretic decision-making under uncertainty.} We answer this by linking the quality of the decisions made directly to the information acquisition channels. In classical bandit theory and partial monitoring, regret have been shown to depend on feedback structure  \citep{lai1985asymptotically,lattimore2020bandit,bartok2014partial}. Side-observation models showed that auxiliary information changes regret rates in a qualitative way \cite{degenne2018bandits}. Recent theoretical work establishes an upper bound on cumulative regret, based on the mutual information between optimal actions and state-action histories \cite{russo2016information}. Moreover, \cite{shufaro2025bits} establishes a clear tradeoff between regret and the information obtained during an action. By showing a sharp bounded-versus-linear phase transition in accumulated regret over time, we provide a unique characterization of the decoder-channel, and demonstrate how cumulative regret is governed by the information a decoder preserves about the true and latent goal.

\vspace{-0.2em}

\section{Limitations and future work}

\emph{Modeling assumptions.} Our analysis assumes that observations are action-independent, excluding active sensing. When agents can act to disambiguate the goal, channels are action-conditioned $C(a)$ hence a policy-induced distribution must take over the bounds. We also exclude temporal dependence, however temporal correlation can be substantial in raw neural signals. With autocorrelated evidence, our effective sample size would shrink. This means our conditional i.i.d.-based bounds will be conservative, instead of being tight.  \emph{Finite-sample measurement.} Proposition~\ref{prop:commit} bounds expected regret, while we measure empirical $\hat R(T)$  from 120 episodes per condition. The envelope satisfaction rates ($85.8\%$ on Stieger2021, $100\%$ on BNCI2014001)  hence reflect the finite-sample fluctuations around the underlying expectation, rather than the failures of the inequality (see Appendix~\ref{app:violation-diagnosis}). Increasing the episode counts would tighten this observed empirical-to-expectation gap. \emph{Scope of experiments.} Our experiments used shared abstract reaching environment to isolate the effect of channels, but further experiments should test whether $\Uid$'s  predictions transfer to deployed control stacks such as wheelchair navigation or prosthetic grasping. Extending analysis beyond goal-committing or hedge-capable planners to belief-space planners gives natural theoretical extension.

\vspace{-0.2em}
\section{Conclusion}
In this paper, we show that decoder evaluation for sequential assistive control should not simply be based on aggregate accuracy, but on the worst-pair Chernoff information $\Gmin$, that is captured in a closed form through the cumulative identification envelope, $\Uid(C)$. In particular, goal-committing planners have an exact bounded-to-linear regret phase transition that is independent of accuracy, namely at $\Gmin = 0$. Our theory states that if all rows of a decoder channel are distinguishable, identification saturates over time, and longer episodes will amortize identification cost. We present $\Uid$ as an actionable diagnostic that can be computed from any confusion matrix, particularly in cases where decoders span orthogonal information channels.

\paragraph{Broader impact.} Our framework is diagnostic, and provides the tools to decide when a decoder is not reliable for intent-aware planning. In settings where safety is critical, such as medical brain-computer interfaces, assistive robotics or autonomous maneuvering with intent prediction, a principled refusal strategy can be valuable~\citep{kim2026guardian}. We note any framework that tries to quantify intent distinguishability could be repurposed for applications in surveillance or attention monitoring, among others. We recommend deployment when the user consent is explicit and revocable, and where channel estimations are safely auditable.

\appendix

\bibliographystyle{plainnat} 
\bibliography{refs}

\appendix

\clearpage
\section{Proof of Theorem~\ref{thm:lower}}
\label{app:lower}

First, note under \textnormal{(A1)}, the latent goal $g^*$ is unseen by the planner, so the only way it influences the planner's information is through the observation sequence $y_{1:t}$. So the information flow is $g^* \rightarrow y_{1:t} \rightarrow \mathcal D_t$, where $\mathcal D_t$ is the processed observations.

Hence, via data-processing inequality,
$I(g^*;\mathcal D_t) \le I(g^*;y_{1:t})$. The information about the true goal in all planner data is at most what is in the raw observation sequence.

We now bound $I(g^*;y_{1:t})$, the \emph{upper bound} on how much information the sequence of observations can have about the true goal.
We use the identity $I(U;V)=H(V)-H(V\mid U)$ to get
\begin{equation}
  I(g^*;y_{1:t})
  =
  H(y_{1:t}) - H(y_{1:t}\mid g^*).
\end{equation}

Since entropy is subadditive, uncertainty in the whole collection must be less than or equal to the sum of uncertainties of all individual pieces. Hence,
\begin{equation}
  H(y_{1:t}) \le \sum_{s=1}^{t} H(y_s).
  \label{eq:subadditivity}
\end{equation}
Under our assumption \textnormal{(A2)} that observations are i.i.d when true goal $g^*$ is fixed, we know the observations are conditionally i.i.d.\ given $g^*$. From independence, $H(y_{1:t}\mid g^*) = \sum_{s=1}^{t} H(y_s\mid g^*)$, and identically distributed gives $ H(y_s | g^*) = H(y_1 | g^*)$. Hence, $H(y_{1:t}\mid g^*) = t\,H(y_1\mid g^*)$ for every $s$.\label{eq:condiid} Combining this with \eqref{eq:subadditivity}, we get
\begin{equation}
  I(g^*;y_{1:t})
  \le
  \sum_{s=1}^{t} H(y_s) - \sum_{s=1}^{t} H(y_s\mid g^*)
  =
  t\,I(g^*;y_1).
  \label{eq:tensor_bound}
\end{equation}

Now we turn to $I(g^*;y_1)$, the information in one time step. Let $E_1\sim\mathrm{Bernoulli}(\pev)$ be the event that an observation arrives at step $1$. Then, $E_1$ is independent of $g^*$, and $y_1$ is equivalent to the pair $(E_1, z_1\mathbf{1}\{E_1=1\})$. Therefore,
\begin{align}
  I(g^*;y_1)
  &=
  I(g^*;E_1) + I(g^*;y_1\mid E_1) \\
  &=
  0 + \pev\,I(g^*;z_1\mid E_1=1) + (1-\pev)\,I(g^*;\bot\mid E_1=0) \\
  &=
  \pev\,\Igz.
  \label{eq:one_step_compute}
\end{align}

This gives our main information-growth bound:
\begin{equation}
  I(g^*;\mathcal D_t)
  \le
  I(g^*;y_{1:t})
  \le
  t\,\pev\,\Igz.
  \label{eq:info_growth}
\end{equation}

Applying Fano's inequality to any estimator $\hat g_t$ measurable with respect to $\mathcal D_t$ gives
\begin{equation}
  P(\hat g_t\neq g^*)
  \ge
  1-\frac{I(g^*;\mathcal D_t)+\log 2}{\log K}
  \ge
  1-\frac{t\,\pev\,\Igz+\log 2}{\log K}.
\end{equation}
Since probability is never negative, a negative lower bound is useless. So we take the positive part, and this proves $P(\hat g_t\neq g^*) \ge \left[1-\frac{t\,\pev\,\Igz + \log 2}{\log K}\right]_+$.

Finally, we sum over $t=1,\ldots,T$ which gives \eqref{eq:M_lower}.
\qed

\clearpage
\section{Proof of Theorem~\ref{thm:upper}}
\label{app:upper}
This time, we show that if a decoder rows are pair-wise distinguishable, the maximum a posteriori (MAP) estimator has error that decays exponentially. 

First, we fix the true goal to be $g^*=i$.
Under the assumptions A1-A4, the observations $y_1,\ldots,y_t$ are i.i.d.\ from the erasure-augmented distribution $\widetilde C_i$. This means that each observation is drawn \emph{independently} from the same erasure-augmented distribution.
Now let $\hat g_t^{\mathrm{MAP}}$ be the MAP rule under the uniform prior. Since the prior is uniform, MAP just selects the goal that has the highest likelihood.

Now consider any competing hypothesis $j\neq i$. The event $\{\hat g_t^{\mathrm{MAP}}=j\}$ would imply$
\widetilde C_j(y_{1:t}) \ge \widetilde C_i(y_{1:t})$, since the likelihood of what MAP chooses must be \emph{at least} as large as the likelihood of the true class. And equivalently, we can rewrite this as
\begin{equation}
  \left(
    \frac{\widetilde C_j(y_{1:t})}{\widetilde C_i(y_{1:t})}
  \right)^s
  \ge 1
  \qquad \forall s\in[0,1].
\end{equation}

Therefore, using Markov's inequality, we get
\begin{align}
  P_i(\hat g_t^{\mathrm{MAP}}=j)
  &\le
  \mathbb{E}_i\!\left[
    \left(
      \frac{\widetilde C_j(y_{1:t})}{\widetilde C_i(y_{1:t})}
    \right)^s
  \right] \\
  &=
  \prod_{\tau=1}^{t}
  \sum_{y\in\{\bot,1,\ldots,K\}}
  \widetilde C_i(y)^{1-s}\widetilde C_j(y)^s \\
  &=
  \left(
    \sum_{y\in\{\bot,1,\ldots,K\}}
    \widetilde C_i(y)^{1-s}\widetilde C_j(y)^s
  \right)^t,
\end{align}
where $P_i$ and $\mathbb{E}_i$ are the probability and expectation under assumption $g^* = i$.

Minimizing over $s\in[0,1]$ we use the definition of Chernoff information gives the bound $P_i(\hat g_t^{\mathrm{MAP}}=j) \le e^{-t\Gij}$. Now we can union-bounding over all wrong hypotheses, which gives $P_i(\hat g_t^{\mathrm{MAP}}\neq i) \le \sum_{j\neq i} e^{-t\Gij}$.
Then, we can average this over the uniform prior to get:
\begin{equation}
\begin{aligned}
P(\hat g_t^{\mathrm{MAP}}\neq g^*)
&\le \frac{1}{K}\sum_{i=1}^{K}\sum_{j\neq i} e^{-t\Gij} \\
&\le (K-1)e^{-t\Gmin}.
\end{aligned}
\end{equation}

We can then sum over $t=1,\ldots,T$, to get
\begin{align}
  \Mcal_{\mathrm{MAP}}(T)
  &\le
  \frac{1}{K}
  \sum_{i=1}^{K}
  \sum_{j\neq i}
  \sum_{t=1}^{T} e^{-t\Gij} \\
  &\le
  \frac{1}{K}
  \sum_{i=1}^{K}
  \sum_{j\neq i}
  \frac{1}{e^{\Gij}-1} \\
  &= \Uid(C) \\
  &\le
  \frac{K-1}{e^{\Gmin}-1}.
\end{align}

This proves Theorem~\ref{thm:upper} when $\Gij>0$. If $\Gij=0$, the finite upper bound is not too meaningful.
\qed

\newpage
\section{Proof of Proposition~\ref{prop:commit} (The regret sandwich)}
\label{app:commit}

This proof applies to goal-committing planners. Goal-committing planners execute an exact policy exactly $\pi^*_{\tilde{g}_t}$ at time $t$. Conditioning on the true goal $g^*=i$, the instantaneous regret equals zero when $\tilde{g}_t = i$. Otherwise, it equals $\delta_{i \tilde{g}_t}$. Therefore, we can write 
$$
\mathbb{E}[c_t^{\mathrm{plan}} - c_t^{\mathrm{oracle}}] = \frac{1}{K} \sum_{i=1}^{K} \sum_{j \neq i} \delta_{ij} P(\tilde{g}_t = j \mid g^* = i).
$$
Assuming A5-C, we have that $\Dlb \leq \delta_{ij} \leq \Dub$. This gives 
\begin{align}
\Dlb \cdot P(\tilde{g}_t \neq g^*)
&\leq \mathbb{E}[c_t^{\mathrm{plan}} - c_t^{\mathrm{oracle}}] \\
&\leq \Dub \cdot P(\tilde{g}_t \neq g^*).
\end{align}

Finally, we sum this over $t=1,2, \cdots, T$ to get the per-step and cumulative bounds (see  \eqref{eq:regret-sandwich}). Then we can substitute $\tilde{g}_t = \hat{g}_t^{\mathrm{MAP}}$, and apply Theorem~\ref{thm:upper} to the upper side. This gives the desired bound $R(T) \leq \Dub \cdot \Uid(C)$. \qed

\vspace{1.5em}
\section{Frozen evaluation protocol}
\label{app:protocol}

Algorithm~\hyperref[app:protocol]{1} details the frozen protocol we use for prediction on BCI datasets. 
\vspace{1em}

\begin{algorithm}[h]
\caption{Frozen prediction protocol.}
\label{alg:protocol}
\begin{algorithmic}[1]
\Require Held-out evaluation data, $K$, $\pev$, environment geometry
\State Estimate $\hat{C}$ on validation split (Laplace smoothing $\lambda = 1$)
\State Compute $\Gij$ via~\eqref{eq:gamma_def}, then $\Gmin$ and $\Uid(\hat{C})$ via~\eqref{eq:Uid_def}
\State Determine $\Dub$ from environment geometry (worst-case wrong-goal cost gap; $\Dub = 2.0$ for $K{=}4$ goals on the unit circle)
\State Set $\Ureg(\hat{C}) \gets \Dub \cdot \Uid(\hat{C})$
\State Freeze all quantities; compute cumulative regret $R(T)$ on held-out evaluation
\State Compare $\Uid$ and $\alpha$ as regret predictors on held-out data only
\end{algorithmic}
\end{algorithm}

\newpage
\section{Hedge-capable planners}
\label{app:hedge}

As opposed to goal-committing planners, there exists a planner class that does not wholly commit to one goal label at each step. Instead, they can take actions that hedge across all plausible goals. Examples include maximum-entropy belief space MPC, BS-MPPI, or Thompson belief planners. For this class of planners, an additional assumption on excess cost-to-misidentification link is required to transfer the identification bounds to regret: 

\textbf{A6-U}.\label{assumption:6-u} There exist some $\Delta_{ub} < \infty$ and non-negative residuals $\eta_t$ with $\sum_{t=1}^{\infty} \eta_t \leq \Cexec$ s.t.
$$
\mathbb{E}[c_t^{\mathrm{plan}} - c_t^{\mathrm{oracle}}] \leq \Dub \cdot P(\hat{g}_t^{\mathrm{MAP}} \neq g^*) + \eta_t.
$$

This assumption states that misidentifying the goal would dominate excess cost, up to some bounded execution residual.
\vspace{0.5em}

\begin{corollary}[]
\label{cor:hedge}
Under assumptions A1-4 and A6-U and Theorem~\ref{thm:upper}, 
\begin{align}
R(T)
&\leq \Dub \cdot \Mcal_{\mathrm{MAP}}(T) + \Cexec \\
&\leq \Dub \cdot \Uid(C) + \Cexec \\
&\leq \Dub \cdot \frac{K-1}{e^{\Gmin} - 1} + \Cexec.
\end{align}
\end{corollary}

\textbf{Empirical validation of A6-U.} We test the assumption directly on Stieger2021. First, we calibrate the constants $\Dub^{\text{hedge}} = 4.80$ and $\Cexec = 4.96$ on the 15-subject calibration cohort. Then, we evaluate on 46 held-out subjects. 

Results showed that $95.2\%$ of cumulative residuals fall within $\Cexec$. On the BNCI2014001 dataset, when constants are refit on two calibration subjects and evaluated on the remaining seven ($\Dub^{\text{hedge}} = 4.75$, $\Cexec = 1.82$), $78.6\%$ fall within $\Cexec$.

Therefore, the assumption A6-U holds approximately for hedge-capable planners on real EEG data. We restrict the exact and assumption-free phase transition to goal-committing planners (Proposition~\ref{prop:commit}), however.

\vspace{1em}
\section{Diagnosing bound violations on Stieger2021}
\label{app:violation-diagnosis}

Proposition~\ref{prop:commit} bounds expected cumulative regret. However, in our experiments we measure empirical  $\hat R(T)$ from $120$ episodes per condition. Even if the underlying expectation respects the bound clearly, we note there are two distinct, finite-sample phenomena that can produce empirical violations:
\begin{enumerate}
    \item The sample variance in $\hat{R}(T)$, and
    \item Estimation noise in $\hat{C}$ that exist at a small validation sample size.
\end{enumerate}

Notably, $14.2\%$ ($227/1{,}594$) of data points violated bounds on Stieger2021. We characterize where these violations cluster, and confirm that both mechanisms are at play. 

\paragraph{Severity distribution.} Out of 227 Stieger violations the median overshoot is $2.0\times$. Around half ($49\%$) overshoot the bound by less than  $2\times$, a third ($32\%$) by $2$-$5\times$, and the remaining $19\%$ by $5\times$ or more. Crucially, no condition shows regret that grows without limit, which would indicate a true $\Gmin = 0$ violation of the Theorem~\ref{thm:upper}.

\emph{Where are violations concentrated?} Perhaps counterintuitively, extreme violations are concentrated at \emph{large} $\Gmin$ where the bound is small in absolute terms, rather than small $\Gmin$ where the theory would predict failure. The reason is finite-sample noise. If the planner commits to an incorrect goal in an unlucky episode, it lifts the episode average over the line when the bound is small. In a loose bound, this same episode does not show.

\vspace{1em}
\section{Extended experimental details}
\subsection{Data}

\textbf{BCI}. Data loaded using MOABB~\citep{jayaram2018moabb,aristimunha2023moabb}.
\begin{itemize}
    \item \textbf{BNCI2014001}~\citep{tangermann2012review}: cross-session evaluation (session 1 train + validation with chronological 80/20 split. session 2 held-out evaluation).
    \item \textbf{Stieger2021}~\citep{stieger2021benefits}: rolling-origin protocol, sessions $1,\cdots, t-2$ for training, session $t-1$ for validation, session $t$ as held-out evaluation. Confusion matrices estimated from hard predictions, with Laplace smoothing, $\lambda=1$. No calibrated posteriors used. 
    Matched-accuracy direction test uses $\delta_\alpha=0.015$, for 417 within-subject pairs across 46 held-out subjects.
\end{itemize}

\textbf{CIFAR-100}. CIFAR-100 via Torchvision~\citep{krizhevsky2009learning,marcel2010torchvision}. 10K train, subsampled from 50K. 10K test, K=100. Feature extraction uses PCA at \{20, 50, 100\} dims, color histograms at \{8, 16\} bins per channel, gradient magnitude in 4$\times$4 spatial grid with 8-bin histograms, spatial mean color at \{4$\times$4, 8$\times$8\} grids, Random Forest and KNN on raw pixels, and calibrated LinearSVC on PCA features. 
\begin{itemize}
    \item Confusion matrices from ground-truth test labels with Laplace smoothing ($\lambda{=}1$). 
    \item Theory quantities computed at $\pev{=}0.2$. 
    \item A commit planner is used for planning, with $T{=}300$, $\pev{=}0.2$, 200~episodes.
    \item  Matched-accuracy threshold is $\delta_\alpha=0.015$ (39 pairs).
\end{itemize}

\textbf{NinaPro multimodal}. NinaPro DB2~\citep{atzori2014electromyography}. 10 subjects, Exercise 2, K=23 grasp movements, 6 repetitions per movement. 4 modalities: EMG (12 ch), accelerometer (36 ch), data glove (22ch), inclinometer (2ch). Features: mean + std per channel. 
\begin{itemize}
    \item Decoder configurations: 14 = 4 modalities $\times$ PCA at multiple dimensions (GLOVE: \{2, 3, 5, 10, full\}, ACC: \{2, 5, 10, full\}, EMG: \{2, 5, full\}, INCLIN: \{2, full\}). 
    \item Cross-validation: leave-one-repetition-out (6-fold). 
    Laplace smoothing $\lambda{=}0.1$ (reduced from 1.0 because with $K{=}23$ and 6 test samples per class, $\lambda{=}1$ would contribute ${\sim}92\%$ of off-diagonal CM mass).
    \item Theory is at $\pev{=}1.0$, and planning is at $\pev{=}0.5$, $T{=}300$, 200~episodes. Matched-accuracy threshold is $\delta_\alpha{=}0.05$, cross-modality, within-subject pairs, subjects as independent unit ($N{=}10$). 
    \item The NinaPro EMG-only negative control uses 11 same-feature decoders (LR, PCA+LR, RP+LR, RF, KNN) on 40 subjects with StratifiedShuffleSplit 75/25 and $\lambda{=}1.0$.
\end{itemize}

\subsection{Setup}
\textbf{Evaluation split}. All theorem-facing constants used in $\Ureg(\hat C)$ must be frozen on a split that is disjoint from the final evaluation split. This includes $\Dub$, $\Cexec$, and any environment-specific geometric constants that are not analytically known. Confusion matrices are estimated on the validation split with Laplace smoothing ($\lambda{=}1$). All theorem-facing quantities ($\Gij$, $\Uid$) are computed from validation CMs and frozen before evaluation. For A6-U calibration, $\Dub$ is set to the 95th percentile of per-step excess-cost-to-misidentification ratios on a subject-disjoint calibration cohort (25\% of subjects), and $\Cexec$ is the 95th percentile of per-condition cumulative residuals on the same cohort.

\textbf{Synthetic environment}. A double-integrator with state $[x,y,v_x,v_y]$, damping~0.1, and $dt{=}0.05$s. Per-step cost is $c_t = \|x_t - x^*(g^*)\|_2$ (Section~\ref{sec:notations} cost with $\lambda=0$). Four goals are at unit distance from the origin, equally spaced angularly. Proportional control with gain~2.0, clipped to $[-1,1]$. For BCI datasets, the planning horizon is $T{=}600$, $\pev{=}1.0$, with 120~episodes per condition.

\textbf{Statistical reporting}. Bootstrap: $B{=}2000$ resamples, Table~\ref{tab:regression} CIs resample conditions, per-subject direction tests resample subjects. All CIs are 95\%. Model-comparison metrics: in-sample OLS $R^2$ with partial-$F$ tests and Spearman $\rho$. Seed for all random number generators is 42.

\textbf{Compute}. All experiments ran on CPU only. Stieger2021: 16 array tasks, each 8 CPUs $\times$ 64\,GB, ~25--90 minutes for each task. BNCI2014001: 8 CPUs $\times$ 16\,GB.

\vspace{1em}
\section{NinaPro per-subject direction accuracy}
\label{app:ninapro-persubject}

See Table~\ref{tab:direction_test} for results on matched-accuracy direction tests on all tested NinaPro subjects.

\section{Controlled phase transition experiment}
\label{app:controlled}

We construct five 4-class confusion matrices each with accuracy $\alpha=0.50$ by fixing the diagonal then redistributing off-diagonal mass, so that one class pair becomes increasingly indistinguishable. This isolates the mechanism behind Corollary~\ref{cor:phase-transition}. With $\pev=0.5$, $T=600$, $100$ episodes per channel (see \emph{(Left)} in Figure~\ref{fig:headline}), $\Gmin$ drops from $4.57\times10^{-2}$ to $7.1\times10^{-5}$. Cumulative regret increases from $22.7$ to $162.6$, a  $7.15\times$ increase.  $\Dub\Uid(C)$ bounds every channel.  This shows where regret variation is not visible to accuracy,   $\Gmin$ correctly tracks it.

\begin{table}[t]
\centering
\caption{Cross-modality matched-accuracy direction test on NinaPro multimodal, with 10 subjects and 70 pairs. Accuracies are within $0.05$, and $\Uid=\infty$ conditions are excluded.  All subjects achieve $\mathcal{U}_{\mathrm{id}}$ direction accuracy above chance. Acc. means direction accuracy, measured as the percentage of correct pairs (the decoder with lower $\Uid$ (resp. higher $\Gmin$) has lower regret).}
\vspace{2em}
\label{tab:direction_test}
\small
\setlength{\tabcolsep}{5pt}
\renewcommand{\arraystretch}{1.1}
\begin{tabular}{lccccc}
\toprule
& & \multicolumn{2}{c}{$\Uid$} & \multicolumn{2}{c}{$\Gmin$} \\
\cmidrule(lr){3-4} \cmidrule(lr){5-6}
Subject & $N$ & Correct & Acc.\ (\%) & Correct & Acc.\ (\%) \\
\midrule
S1  & 6  & 5/6  & 83 & 6/6  & 100 \\
S2  & 5 & 3/5 & 60 & 4/5 & 80 \\
S3  & 6 & 5/6 & 83 & 5/6 & 83 \\
S4  & 9  & 6/9  & 67 & 5/9  & 56 \\
S5  & 7 & 5/7 & 71 & 4/7 & 57 \\
S6  & 7  & 4/7  & 57 & 4/7  & 57 \\
S7  & 9  & 8/9  & 89 & 7/9  & 78 \\
S8  & 7 & 5/7 & 71 & 4/7 & 57 \\
S9  & 8 & 7/8 & 88 & 4/8 & 50 \\
S10 & 6  & 6/6  & 100 & 4/6 & 67 \\
\midrule
Mean & 7.0 & --- & 77 & --- & 68 \\
\bottomrule
\end{tabular}
\end{table}

\end{document}